\documentclass{article}
\usepackage{spconf,amsmath,graphicx,hyperref}
\usepackage{float}
\usepackage{cite}
\usepackage{spconf,amsmath,graphicx,hyperref}
\usepackage{multirow, multicol}
\usepackage{cite}

\title{Bridging the Gap: A Comparative Exploration of Speech-LLM and End-to-End Architecture for Multilingual Conversational ASR}
\name{Yuxiang Mei$^{1}$, Dongxing Xu$^{2}$, Jiaen Liang$^{2}$, Yanhua  Long$^{1}$\sthanks{Yanhua Long is the corresponding author. This work was sponsored by Natural Science Foundation of Shanghai (Grant No.25ZR1401277).}}
\address{
  $^1$Shanghai Normal University, Shanghai, China\\
  $^2$Unisound AI Technology Co., Ltd. Beijing, China
  }

\begin{document}
%
\maketitle
\begin{abstract}
The INTERSPEECH 2025 Challenge on Multilingual Conversational Speech Language 
Models (MLC-SLM) promotes multilingual conversational ASR with large language models (LLMs). 
Our previous SHNU-mASR system adopted a competitive parallel-speech-encoder architecture 
that integrated Whisper and mHuBERT with an LLM. However, it faced two challenges: 
simple feature concatenation may not fully exploit complementary information, 
and the performance gap between LLM-based ASR and end-to-end (E2E) encoder–decoder ASR remained unexplored.
In this work, we present an enhanced LLM-based ASR framework that combines fine-tuned Whisper 
and mHuBERT encoders with an LLM to enrich speech representations. We first evaluate E2E Whisper 
models with LoRA and full fine-tuning on the MLC-SLM ASR task, and then propose 
cross-attention-based fusion mechanisms for the parallel-speech-encoder. On the official evaluation set of the MLC-SLM Challenge, 
our system achieves a CER/WER of 10.69\%, ranking on par with the top-ranked Track 1 systems, even though it uses only 1,500 hours of baseline training data compared with their large-scale training sets. Nonetheless, we find that our final LLM-based ASR still does not match the performance of a fine-tuned E2E Whisper model, providing valuable empirical guidance for future Speech-LLM design. Our code is publicly available at \url{https://github.com/1535176727/MLC-SLM}.

\end{abstract}
\begin{keywords}
ASR, LLM, multilingual, conversational
\end{keywords}

\section{Introduction}
\label{sec:intro}

In recent years, the integration of large-scale pre-trained speech models (e.g., Whisper \cite{radford2023robust}) with large language models (LLMs) \cite{bai2023qwen, gonzalez2025salamandra,touvron2023llama, achiam2023gpt, brown2020language} has attracted widespread attention.  Integrating powerful speech encoders with LLMs has demonstrated great potential in improving multilingual and conversational ASR by providing richer linguistic modeling and stronger cross-lingual generalization \cite{tang2023salmonn, chen2025minmo, xue2024ideal, song2024comparative, bai2024seed}. In this context, the INTERSPEECH 2025 Multilingual Conversational Speech Language Modeling (MLC-SLM) Challenge\footnote{\url{https://www.nexdata.ai/competition/mlc-slm}} investigates how LLMs can be leveraged to advance multilingual conversational ASR on real-world data.

Existing approaches in the MLC-SLM Challenge include encoder–decoder integration, 
where speech features are projected into LLM embedding space \cite{concina2025eloquence, meng2025ilt}, 
prompt-based modeling using language-specific instructions \cite{peng2025ntu, li2025seewo, peng2025bi}, and 
parallel-speech-encoder frameworks that fuse complementary representations from multiple pre-trained encoders \cite{xue2025tea, lin2025diarization, mei25_mlcslm}. Our previous work \cite{mei25_mlcslm}, SHNU-mASR, belongs to the last category, combining Whisper and 
mHuBERT \cite{boito2024mhubert} encoders in parallel with an LLM, and despite being trained only on the 1,500-hour baseline dataset, 
it achieved highly competitive results among numerous challenge systems that relied on large-scale training data.

Despite these advances, existing methods face two key limitations. First, most fusion strategies rely on simple 
feature concatenation, which may not fully exploit the complementary information across heterogeneous encoders. 
Second, Speech-LLM systems have rarely been compared with carefully fine-tuned end-to-end (E2E) ASR systems such as 
Whisper under the same training data conditions, leaving the actual performance gap unclear. 
To address these challenges, this paper presents a systematic study with the following contributions:
\begin{itemize}
\item Analyze the performance of Whisper encoders under both LoRA-based~\cite{hu2022lora} and 
full fine-tuning strategies on the MLC-SLM ASR task.
\item Propose and compare several cross-attention-based fusion mechanisms 
for parallel-speech-encoder architecture.
\item Investigate the impact of different projector designs (linear vs. Q-Former~\cite{li2023blip2}) 
between speech encoders and LLM on MLC-SLM ASR performance.
\item Provide direct comparisons between our Speech-LLM system and well fine-tuned E2E Whisper models, 
offering empirical insights for future Speech-LLM design.
\end{itemize}

\begin{figure*}
    \centering
    \includegraphics[width=0.9\linewidth]{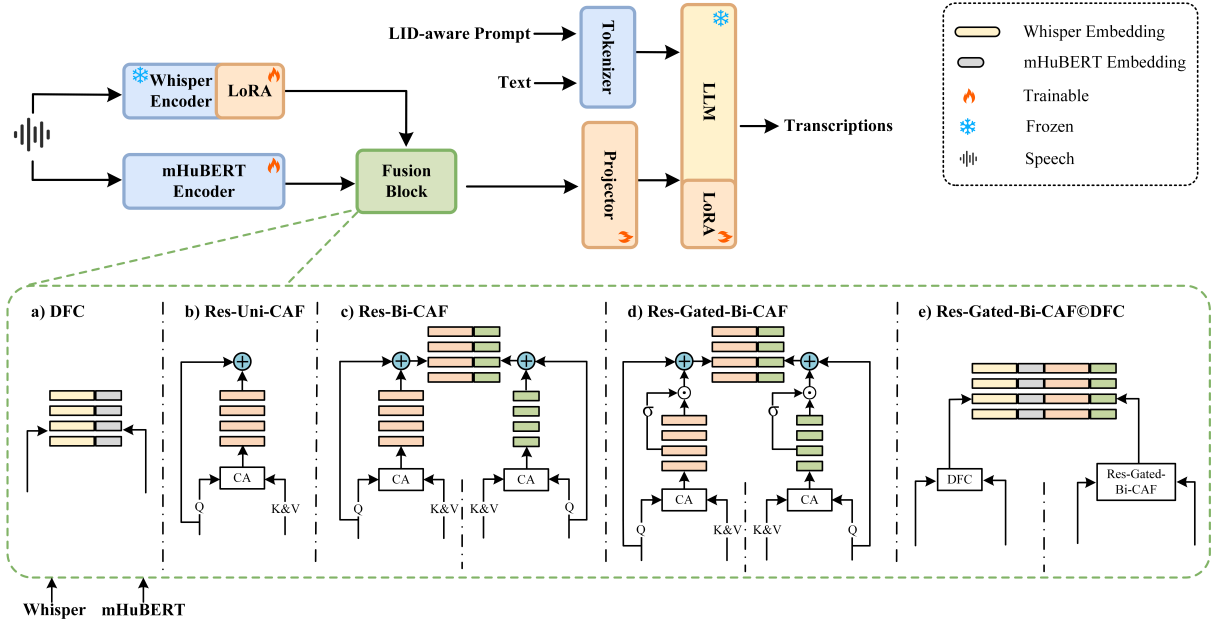}
    \caption{Overall architecture with different speech representation fusion mechanisms. }
    \label{fig:model}
    \vspace{-0.3cm}
\setlength{\abovecaptionskip}{-0.3cm}
\end{figure*}

\vspace{-0.3cm}
\section{Fine-Tuning Whisper for MLC-SLM ASR}
\label{sec:format}

Large-scale pre-trained speech models such as Whisper Large-v3 \cite{radford2023robust} represent 
the current state-of-the-art in end-to-end (E2E) multilingual ASR. Trained on a mixture of approximately 
1 million hours of weakly labeled speech and 4 million hours of pseudo-labeled speech generated using Whisper-large-v2, 
Whisper Large-v3 demonstrates strong cross-lingual robustness and zero-shot generalization. 
In the context of the MLC-SLM Challenge, Whisper serves as a strong baseline system for multilingual 
conversational ASR. However, directly applying the frozen model often leaves room for improvement 
due to domain mismatch and conversational speech variability. This motivates the need for 
task-specific adaptation through parameter-efficient finetuning methods such as LoRA \cite{hu2022lora}, 
as well as full-model finetuning.

In this work, we investigate both LoRA-based and full finetuning strategies on Whisper Large-v3 
for the MLC-SLM ASR task. Specifically, LoRA modules are inserted into the query and value projections 
of self-attention layers, enabling efficient adaptation with limited trainable parameters. 
In parallel, we explore full finetuning to assess the upper-bound performance when all parameters are updated. 
These experiments serve \textbf{two purposes}: 1) establish a strong E2E baseline for fair comparison 
with Speech-LLM architectures, and 2) provide insights into how parameter-efficient 
adaptation compares with full optimization in a multilingual conversational ASR. 
The findings from this section form the basis for the proposed heterogeneous encoder fusion-based 
Speech-LLM systems.

\vspace{-0.1cm}
\section{Proposed Methods}
\label{sec:pagestyle}

\subsection{Architecture}
\label{ssec:subhead}

The overall structure of our proposed architecture is illustrated in Fig.~\ref{fig:model}. 
It follows a parallel-speech-encoder design, where the Whisper encoder 
and the mHuBERT encoder process the input speech in parallel. 
The resulting representations are then fused using one of several proposed 
fusion mechanisms, followed by a linear projector to match the embedding space of 
the large language model (LLM). Finally, the fused embeddings are fed into the LLM 
for transcription generation.  
Compared with our previous SHNU-mASR \cite{mei25_mlcslm} system submitted to the 2025 MLC-SLM Challenge, 
the fusion stage is refined with more explicit 
design choices, as highlighted in the lower part of Fig.~\ref{fig:model}.
In addition, both linear projection and Q-Former~\cite{li2023blip2} are explored to enhance 
cross-modal alignment and better leverage complementary information from different encoders. 
The details of these design choices are presented in the following sections.

\subsection{Parallel-Speech-Encoder Fusion Mechanism}
\label{ssec:Mechanism}
To investigate effective strategies for combining parallel speech encoders, we explore multiple fusion 
mechanisms that integrate Whisper representations $h^w \in \mathbb{R}^{T \times d_w}$ and mHuBERT 
representations $h^m \in \mathbb{R}^{T \times d_m}$. As demonstrated in the lower 
part of 
Fig.~\ref{fig:model}, five fusion mechanisms are explored and detailed as below: 
\begin{itemize}
    \item \textbf{Direct Feature Concatenation (DFC).}  
    As a simple yet effective baseline, DFC directly concatenates the hidden representations from both encoders:
    \begin{equation}
    h^{\text{DFC}} = \text{Concat}(h^w, h^m) 
    \end{equation}
where $h^{\text{DFC}}$ preserves all encoder information without additional interaction, providing a straightforward joint representation.  

    \item \textbf{Unidirectional Cross-Attention Fusion with Residual Connection (Res-Uni-CAF).}  
    In this mechanism, Whisper embeddings serve as queries to attend to mHuBERT features through cross-attention:
    \begin{equation}
    h^{w \leftarrow m} = \mathrm{CrossAttn}(Q = h^w, K=V = h^m)
    \end{equation}
    The attended features are then integrated into the original Whisper embeddings via a residual connection:
    \begin{equation}
    h^{\text{Res-Uni-CAF}} = h^{w \leftarrow m} + h^w 
    \end{equation}
    This design enables Whisper to selectively enrich its representations with complementary mHuBERT information while maintaining its original features.  

    \item \textbf{Bidirectional Cross-Attention Fusion with Residual Connection (Res-Bi-CAF).}  
    Extending Res-Uni-CAF, this mechanism employs bidirectional interaction between the two encoders. Whisper attends to mHuBERT as in Res-Uni-CAF, while mHuBERT also attends to Whisper:
    \begin{equation}
    h^{m \leftarrow w} = \mathrm{CrossAttn}(Q = h^m, K=V = h^w)
    \end{equation}
    The final representation is obtained by concatenating both residual-enhanced embeddings:
    \begin{equation}
    h^{\text{Res-Bi-CAF}} = \text{Concat}( h^{w \leftarrow m} + h^w, \; h^{m \leftarrow w} + h^m)
    \end{equation}
    This bidirectional scheme captures richer mutual dependencies between the two encoders.  

\item \textbf{Gated Res-Bi-CAF (Res-Gated-Bi-CAF).}  
Building on Res-Bi-CAF, this enhanced mechanism introduces learnable gating to control the contribution of attended features. For Whisper, the integration is defined as:
\begin{equation}
\tilde{h}^w = \sigma(W_{g}^{w \leftarrow m} h^{w \leftarrow m}) \odot h^{w \leftarrow m} + h^w
\end{equation}
while for mHuBERT, the process is symmetric:
\begin{equation}
\tilde{h}^m = \sigma(W_{g}^{m \leftarrow w} h^{m \leftarrow w}) \odot h^{m \leftarrow w} + h^m
\end{equation}
The final fused representation concatenates both gated outputs:
\begin{equation}
h^{\text{Res-Gated-Bi-CAF}} = \text{Concat}(\tilde{h}^w, \tilde{h}^m)
\end{equation}
Here, $\sigma(\cdot)$ denotes the element-wise sigmoid activation that outputs values in $(0,1)$ to adaptively regulate the contribution of attended features, $W_{g}^{w \leftarrow m}$ and $W_{g}^{m \leftarrow w}$ are learnable gating matrices, and $\odot$ represents element-wise multiplication. The gating mechanism allows dynamic regulation of encoder contributions, offering more flexible and adaptive fusion.  

\item \textbf{Res-Gated-Bi-CAF with DFC (Res-Gated-Bi-CAF©DFC).}  
This hybrid approach combines residual gated bi-directional cross-attention fusion with direct feature concatenation. The fused representation is obtained as:
\begin{equation}
h^{\text{Res-Gated-Bi-CAF©DFC}} = \text{Concat}(h^{\text{DFC}}, h^{\text{Res-Gated-Bi-CAF}}).
\end{equation}
This design leverages both raw fused features and adaptively integrated features.  
\end{itemize}

\subsection{Projector Structure}
\label{ssec:projector}

To find the most effective bridge between speech encoders and the LLM,  
two projectors are investigated in this study:  1) \textbf{Linear Projector.} It is our baseline projector from \cite{mei25_mlcslm}, which uses a series of 1D convolutions and an MLP to perform temporal downsampling and dimension mapping; 2) \textbf{Q-Former Projector.} As an alternative, we evaluated a Q-Former \cite{li2023blip2}, which uses a set of learnable queries to summarize the fused speech features.


\subsection{Training Strategy}
\label{ssec:strategy}

In our previous work SHNU-mASR~\cite{mei25_mlcslm}, we adopted a tri-stage training strategy within the speech-LLM 
framework. Specifically, the projector was first trained to connect speech encoder and LLM, then the  
parallel-speech-encoder was adapted with a frozen LLM, and finally the LLM was adapted jointly with the 
projector and speech encoders. Motivated by the need for both efficiency and stable convergence in large-scale 
multilingual ASR, we instead employ a streamlined two-stage paradigm. We first fine-tune the speech encoders 
independently: Whisper is fine-tuned using LoRA~\cite{hu2022lora} or full-parameter strategies, while mHuBERT is 
adapted through CTC-based~\cite{graves2012connectionist} fine-tuning. After this step, the tuned encoders are 
connected to the LLM. In the first stage of joint training, only the projector is optimized. In the second stage, we 
update both the projector and the LLM parameters via LoRA, while keeping the speech encoders fixed.

\vspace{-0.2cm}
\section{EXPERIMENT SETUP}

\label{sec:typestyle}

\subsection{Datasets}
\label{ssec:datastes}
The experiments are conducted on the 1500-hour multilingual conversational speech dataset provided by the MLC-SLM Challenge , which covers 11 languages. To ensure a fair evaluation and strict data separation, we adopt the following setup:
\begin{itemize}
    \item \textbf{Validation-Set (Valid) :} A new internal development set is constructed by randomly sampling approximately 2 hours of speech per language from the official training set. This set is used for model selection and early stopping.
    \item \textbf{Development-Set (Dev) :} The official development set provided by the challenge is reserved exclusively for system evaluation.
    \item \textbf{Evaluation-Set (Eval) :} The official evaluation set provided by the challenge is reserved exclusively for final performance reporting and leaderboard submission.
    \item \textbf{OOD CV-TestSet (CV-Test) :} To evaluate out-of-domain robustness, we additionally sample 2 hours of test speech per language from the CommonVoice 21.0 dataset. Since the test data for Korean (0.6 h) and Vietnamese (1.4 h) are shorter than 2 hours, we directly adopt their original test sets.
\end{itemize}

\subsection{Configurations}
\label{ssec:modelconf}

The LoRA fine-tuned Whisper-Large-v3 encoder uses rank $r=32$ and $\alpha=64$, 
while the mHuBERT-147 model is fully fine-tuned with a CTC objective. 
Qwen2.5-7B~\cite{yang2024qwen2.5} serves as the LLM. Training employs 
the Adam optimizer with a learning rate of $1\times10^{-4}$, linear decay 
with 4000 warmup steps, weight decay of 0.01, and gradient clipping at norm 5. 
Dynamic batching is applied with up to 120 seconds of audio per batch. 
Models are trained for 6 epochs per stage on four NVIDIA A100 80G GPUs with bfloat16 precision.
As an E2E ASR baseline, Whisper-Large-v3 is fine-tuned on the same data with 
comparable strategies (including LoRA with identical $r$ and $\alpha$) to 
ensure fair comparison with Speech-LLM models.

For evaluation, Character Error Rate (CER) is reported for non-Latin languages (Japanese, Korean, Thai), 
and Word Error Rate (WER) for others, using the meeteval toolkit~\cite{von2023meeteval} 
for consistent multilingual scoring.
\vspace{-0.2cm}
\section{Results and Discussions}
\label{sec:majhead}

All results reported in this section are obtained using greedy search decoding, together with the inference-time Whisper text normalization~\cite{radford2023robust} and 5-gram repetition removal. This procedure removes spurious repetitive tokens from the LLM output, which helps to improve the overall text quality and reduces redundant predictions.

\subsection{Results of E2E ASR Baselines}
\label{ssec:resultswhisper&hubert}

\vspace{-15pt}
\begin{table}[h]

\centering
\caption{WER/CER (\%) of different fine-tuning strategies on Dev-Set, Eval-Set, and OOD CV-TestSet.}
\vspace{5pt}

\label{tab:whisper-finetune}
\begin{tabular}{lccc}
\toprule
\textbf{System} & \textbf{Dev} & \textbf{Eval} & \textbf{CV-Test} \\
\midrule
Vanilla Whisper & 16.04 & 15.96 & \textbf{9.86} \\
Whisper (LoRA-fine-tuned) & 11.40 & 10.71 & 11.47 \\
Whisper (Full-fine-tuned) & \textbf{10.99} & \textbf{10.07} & 13.11 \\
mHuBERT (CTC)$^{*}$ & 29.67 & 25.65 & 68.49 \\
\bottomrule
\end{tabular}

\vspace{0.1cm}
\footnotesize{$^{*}$ The mHuBERT (CTC) system was fine-tuned for only 32 epochs. }
\end{table}

\vspace{-25pt}
\begin{table}[ht]
\centering
\caption{Impact of using different projector structures for parallel-speech-encoders are with direct concatenation feature fusion.}
\vspace{5pt}
\label{tab:proj-enc}
\begin{tabular}{lccc}
\toprule
\textbf{Projector} & \textbf{Dev} & \textbf{Eval} & \textbf{CV-Test} \\
\midrule
Linear & 11.91 & 11.05 & 14.91 \\
Win-level Q-Former & 12.52 & 11.51 & 15.64 \\
\bottomrule
\end{tabular}
\end{table}
\vspace{-5pt}

\begin{table*}[!ht]
\renewcommand\arraystretch{1.2}
\centering
\caption{WER/CER (\%) of Speech-LLM models with different parallel-speech-encoder fusion mechanisms. 
Stage~1: only the projector is trained; 
Stage~2: LLM LoRA and projector are jointly optimized. mHuBERT is fully finetuned with CTC.}
\vspace{5pt}
\label{tab:fusion-llm}
\setlength{\tabcolsep}{3.8pt}
\scalebox{0.9}{
\begin{tabular}{l|ccc|ccc|ccc|ccc}
\hline
\multirow{3}{*}{\textbf{System}} 
& \multicolumn{6}{c|}{\textbf{LoRA finetuned whisper with mHuBERT}} 
& \multicolumn{6}{c}{\textbf{Full finetuned whisper with mHuBERT}} \\
\cline{2-13}
& \multicolumn{3}{c|}{\textbf{Stage 1}} 
& \multicolumn{3}{c|}{\textbf{Stage 2}} 
& \multicolumn{3}{c|}{\textbf{Stage 1}} 
& \multicolumn{3}{c}{\textbf{Stage 2}} \\
\cline{2-13}
& \textbf{Dev} & \textbf{Eval} & \textbf{CV-Test} 
& \textbf{Dev} & \textbf{Eval} & \textbf{CV-Test} 
& \textbf{Dev} & \textbf{Eval} & \textbf{CV-Test} 
& \textbf{Dev} & \textbf{Eval} & \textbf{CV-Test} \\
\hline
DFC & 12.94 & 12.11 & 16.27 & 11.91 & 11.05 & 14.91 & 12.81 & 11.62 & 16.97 & 11.94 & 10.84 & 15.37 \\
Res-Uni-CAF & 13.04 & 12.05 & 16.28 & 12.00 & 11.03 & 14.69 & 12.82 & 11.69 & 17.02 & \textbf{11.74} & \textbf{10.69} & \textbf{15.26} \\
Res-Bi-CAF & 12.91 & 11.90 & 15.83 & 11.97 & 11.10 & 14.69 & 12.84 & 11.53 & 17.14 & 11.94 & 10.81 & 15.47 \\
Res-Gated-Bi-CAF & 12.68 & 11.83 & 15.83 & 11.94 & 11.23 & \textbf{14.56} & \textbf{12.60} & 11.46 & 16.90 & 11.90 & 10.77 & 15.33 \\
Res-Gated-Bi-CAF©DFC & \textbf{12.56} & \textbf{11.57} & \textbf{15.55} & \textbf{11.90} & \textbf{11.02} & \textbf{14.56} & 12.62 & \textbf{11.44} & \textbf{16.53} & 12.05 & 10.90 & 15.54 \\
SHNU-mASR$^{*}$ & 13.76 & 11.83 & 17.49 & 13.39 & 11.43 & 19.86 & - & - & - & - & - & - \\
\hline
\end{tabular}}

\vspace{0.1cm}
\footnotesize{$^{*}$ In SHNU-mASR, Stage~1 means joint training of the speech encoder and projector, while Stage~2 further incorporates the LLM.}
\vspace{-0.3cm}
\end{table*}

From Table~\ref{tab:whisper-finetune}, we observe that full fine-tuning of Whisper 
achieves the best performance on the in-domain sets (Dev and Eval), while LoRA fine-tuning also brings notable improvements over the vanilla baseline. However, on the out-of-domain CV-Test set, vanilla Whisper shows the lowest WER, whereas both fine-tuning 
methods reduce robustness, with full fine-tuning suffering the most. These results reveal 
a clear trade-off between in-domain adaptation and cross-domain generalization in 
multilingual ASR. Furthermore, the large performance gap between mHuBERT and Whisper 
suggests that they may provide complementary information, which can be exploited to 
further enhance speech-LLM models when combined.

\vspace{-0.3cm}
\subsection{Speech-LLM Models with Different Projectors}
\label{ssec:diffprojector}

Table~\ref{tab:proj-enc} reports using different projector designs in the parallel-speech-encoder setting, where a LoRA-finetuned Whisper encoder and a fully fine-tuned mHuBERT 
encoder are fused via direct concatenation. We see that the Linear projector outperforms 
the more complex Win-level Q-Former across all test sets, achieving lower WER/CER on both 
in-domain and out-of-domain data. This indicates that a simple linear mapping is not only 
sufficient for aligning fused speech representations with the LLM input space, but also more 
robust and efficient than the heavier Q-Former design.

\vspace{-0.3cm}
\subsection{Parallel-speech-encoder Fusion Mechanisms with Speech-LLM Architecture}
\label{ssec:fusionmechanisms} 

Table~\ref{tab:fusion-llm} shows the results to examine the impact of different fusion mechanisms for parallel speech encoders. By comparing the first and last rows, we observe that when using an end-to-end finetuned Whisper as the speech encoder within the speech-LLM framework, a simple two-stage training strategy proves more effective on both Dev and CV-Test sets in Stage~1 and achieves better performance across all three test sets in Sgate~2. This approach outperforms the commonly adopted tri-stage scheme, highlighting the importance of carefully designing the training procedure for each component in speech-LLM.

When examining the first five rows of Table~\ref{tab:fusion-llm}, 
we find that in Stage~1, gated fusion consistently outperforms both 
the simple DFC and the proposed Res-Uni(Bi)-CAF, achieving the 
largest improvements on Dev and Eval sets while maintaining strong 
robustness on the OOD CV-Test. However, in Stage~2, these gains 
diminish substantially, suggesting that once the speech encoders 
are sufficiently task-specifically finetuned, the choice of fusion 
mechanism for parallel encoders becomes less critical, and more 
complex fusion methods yield only marginal additional improvements.

Finally, comparing the use of LoRA-finetuned versus fully finetuned 
Whisper encoders, we observe that the fully finetuned model brings 
gains only on the Eval set, while offering almost no benefit on the Dev set and  
causing a significant performance drop on the OOD CV-Test set. 
This indicates that within speech-LLMs, employing a LoRA-finetuned 
speech encoder is  enough and preferable for 
generalization.

\vspace{-0.3cm}
\subsection{Performance of Speech-LLM vs. E2E ASR Systems}
\label{ssec:speechllm_vs_e2e}

Table~\ref{tab:speechllm-e2e} compares our proposed Speech-LLM system with several E2E ASR baselines and top-ranked Speech-LLM submissions in the MLC-SLM challenge. Our model achieves strong performance 
without relying on any external training resources, delivering competitive results even against 
systems such as Seewoo and NTU-Speechlab that used large-scale external datasets. 
Compared with our previous SHNU-mASR, the proposed approach yields substantial improvements on 
both the in-domain and OOD CV-Test sets, demonstrating enhanced robustness and generalization. 
These results highlights the effectiveness of first fine-tuning speech encoders on in-domain data 
before integrating them into the LLM, which proves more effective than the tri-stage pipeline.

\vspace{-0.3cm}
\begin{table}[!htbp]
\centering
\caption{Comparison of Speech-LLM and E2E ASR Systems.}
\vspace{5pt}
\label{tab:speechllm-e2e}
\scalebox{0.9}{
\begin{tabular}{l c c c}
\toprule
\textbf{System} & \textbf{Dev} & \textbf{Eval} & \textbf{CV-Test} \\
\midrule
Whisper (LoRA-fine-tuned) & 11.40 & 10.71 & \textbf{11.47} \\
Whisper (Full-fine-tuned) & \textbf{10.99} & \textbf{10.07} & 13.11 \\
mHuBERT (CTC) & 29.67 & 19.99 & 68.49 \\
\midrule
NTU-Speechlab \cite{peng2025ntu} & 11.57 & 10.58 & - \\
Seewoo\cite{li2025seewo} & 12.73 & 11.57 & - \\
SHNU-mASR\cite{mei25_mlcslm} & 13.39 & 11.43 & 19.86 \\
Proposed Speech-LLM & 11.74 & 10.69 & 15.26 \\
\bottomrule
\end{tabular}}
\vspace{-0.2cm}
\end{table}

Moreover, it is worth noting that, despite these advantages, a performance gap remains 
compared with fully fine-tuned E2E Whisper systems trained on the same data. This 
suggests that current speech-LLMs still face inherent limitations, indicating 
substantial room for improvement in model design and optimization.
\vspace{-0.2cm}
\section{Conclusions }
\label{sec:conclusions}

In this work, we show that fine-tuning the speech encoders before integrating them with the LLM is 
more effective than the tri-stage training pipeline in speech-LLM architecture. 
Although employing complex fusion mechanisms can 
yield additional performance gains during the projector training stage, 
these improvements largely diminish after joint optimization with the LLM. 
Finally, our results reveal that Speech-LLM still exhibits a performance gap 
compared with fully fine-tuned end-to-end Whisper systems trained on the 
same amount of speech data, indicating substantial room for improvement in model 
design and training methodology.

\bibliographystyle{IEEEbib}
\bibliography{strings,refs}

\end{document}